\documentclass[sigconf]{acmart}
\renewcommand\footnotetextcopyrightpermission[1]{} 
\usepackage{stfloats}
\usepackage[bordercolor=white,backgroundcolor=gray!30,linecolor=black,colorinlistoftodos]{todonotes}

\makeatletter
\def\algbackskip{\hskip\dimexpr-\algorithmicindent-\labelsep}
\def\LState{\Statex \algbackskip}
\makeatother

\makeatletter
\def\subsubsection{\@startsection{subsubsection}{3}%
  \z@{.5\linespacing\@plus.7\linespacing}{.1\linespacing}%
  {\normalfont\itshape}}
\makeatother

\renewcommand\footnotetextcopyrightpermission[1]{} 
\usepackage{url}

\usepackage{listings}

\usepackage{xspace}

\usepackage{placeins}

\usepackage{algorithm}
\usepackage[noend]{algpseudocode}
\algnewcommand{\LineComment}[1]{\State \emph{\textcolor{blue}{\(\triangleright\) #1}}}
\algrenewcommand\algorithmicindent{1em}%
\newcommand{\pluseq}{\mathrel{+}=}
\usepackage{graphicx}
\usepackage{caption}
\usepackage{subcaption}
\usepackage{tabularx}
\usepackage{enumitem}   


\usepackage{color, soul} 
\soulregister\emph7

\begin{abstract}
Deep learning (DL) is one of the most prominent branches of machine learning. Due to the immense computational cost of DL workloads, industry and academia have developed DL libraries with highly-specialized kernels for each workload/architecture, leading to numerous, complex code-bases that strive for performance, yet they are hard to maintain and do not generalize. In this work, we introduce the \emph{batch-reduce GEMM kernel} and show how the most popular DL algorithms can be formulated with this kernel as the basic building-block. Consequently, the DL library-development degenerates to mere (potentially automatic) tuning of loops around this sole optimized kernel. By exploiting our new kernel we implement Recurrent Neural Networks, Convolution Neural Networks and Multilayer Perceptron training and inference primitives in just 3K lines of high-level code. Our primitives outperform vendor-optimized libraries on multi-node CPU clusters, and we also provide proof-of-concept CNN kernels targeting GPUs. Finally, we demonstrate that the batch-reduce GEMM kernel within a tensor compiler yields high-performance CNN primitives, further amplifying the viability of our approach.
\end{abstract}

\begin{document}
%

\title{High-Performance Deep Learning via a Single Building Block}

\author{Evangelos Georganas, Kunal Banerjee, Dhiraj Kalamkar, Sasikanth Avancha, Anand Venkat, Michael Anderson, Greg Henry, Hans Pabst, Alexander Heinecke}
\affiliation{\institution{Intel Corporation}}

\renewcommand{\shortauthors}{E. Georganas et al.}

\maketitle

%

\section{Introduction}
In the past decade, machine learning has experienced an academic and industrial renaissance where deep learning (DL) has been the main driving force. More specifically, deep neural networks have advanced the fields of computer vision, speech recognition, machine translation and search ranking, and naturally emerge in numerous applications and scientific domains~\cite{origalexnet,szegedy2015going,simonyan2014very,yu2013feature,wu2016google,cheng2016wide}.

Three types of neural networks (NN) comprise the most prominent DL workloads by representing 95\% of the data-centers's demands~\cite{jouppi2017datacenter}: i) Recurrent Neural Networks (RNN)~\cite{graves2013speech} with the so-called Long Short-Term Memory (LSTM)~\cite{hochreiter1997long} networks being the most popular variation, ii) Convolution Neural Networks (CNN)~\cite{origalexnet}, and iii) Multi-Layer Perceptrons (MLP)~\cite{minsky2017perceptrons,hornik1989multilayer}. Additionally, the contemporary Transformer~\cite{transformer} and BERT~\cite{bert} workloads computationally involve fully-connected layers which also lie in the heart of MLP. All these neural networks can be further associated with two use-cases: \emph{training} of the underlying NN models (i.e.\ learning via back-propagation~\cite{lecun1988theoretical}), and \emph{inference} (i.e.\ yielding predictions) based on trained models. Due to the increase of the involved datasets' size and complexity in deep neural networks (DNN), the training and inference tasks require vast amount of computation. Therefore, academia and industry have invested into the development of DL libraries targeting all the aforementioned workloads on various architectures.

\begin{figure}[t!]
\centering
\includegraphics[width=\columnwidth]{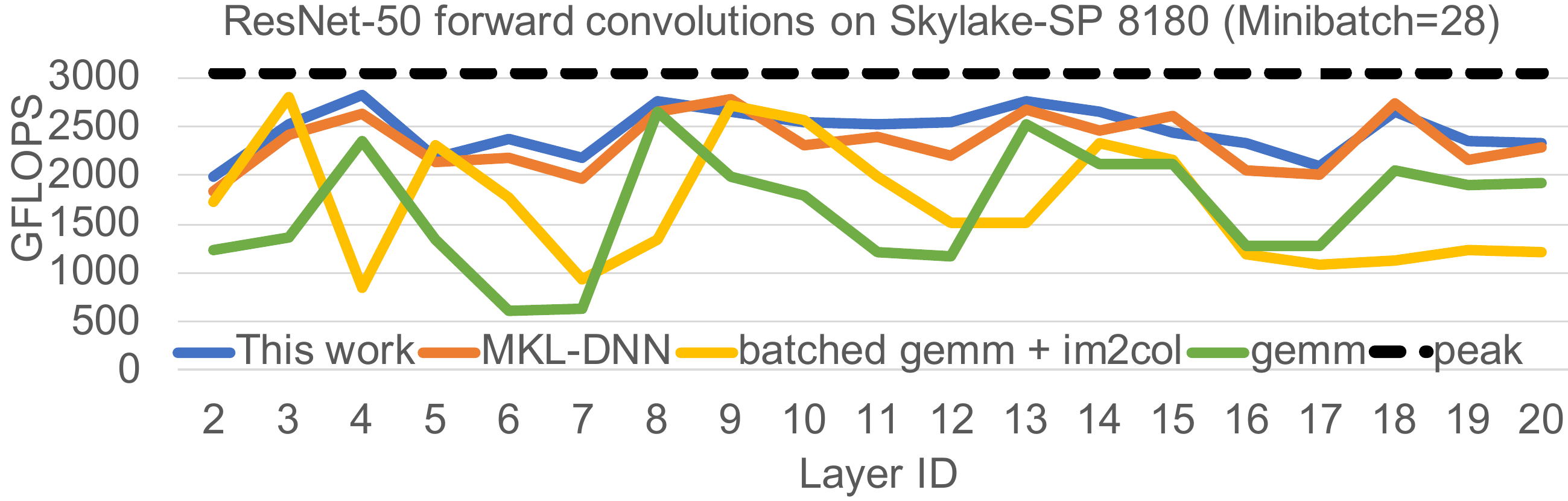}
\caption{Performance of ResNet-50 forward convolutions}
\label{fig:motivation}
\end{figure}

The development of such DL libraries typically embraces one of the following strategies: (i) the specific workload kernel leverages coarse-grained, linear algebra library calls, e.g.\ LSTM cell via large GEneral Matrix Multiply (GEMM) calls in mkl-dnn~\cite{mkldnn}, convolutions via image-to-column tensor transformations and subsequent large GEMM calls~\cite{vasudevan2017parallel,anderson2017low}, or (ii) for each workload and use-case (training/inference) the kernel employs a specialized implementation that targets the specific algorithm/workload and architecture at hand, e.g.\ convolution kernels in mkl-dnn and cuDNN~\cite{chetlur2014cudnn}. The former approach of deploying coarse-grained, linear algebra library calls provides ease in the DL library development process since no special kernel development is involved. However it may result in suboptimal data reuse (e.g.\ redundant data movements to format underlying tensor/matrices in the required layout that enables GEMM calls), and also it is not flexible enough to allow efficient, fine-grained fusion of other operators. The latter approach of implementing specialized kernels for each DL workload/use-case and platform/architecture strives for performance but naturally results in numerous, complex code-bases that are hard to maintain and do not generalize. For example, the code-base \emph{only for convolutions on CPUs} within mkl-dnn consists of $\sim$36,000 lines of code. Figure~\ref{fig:motivation} shows the performance of various convolution kernel implementations on a Xeon Skylake-SP 8180 processor. The yellow and green lines represent implementations adopting strategy (i). More specifically, the green line shows the performance of convolutions that leverage small GEMM library calls, whereas the yellow line illustrates the performance of an implementation which uses image to column transformations and \emph{batched GEMM}~\cite{dongarra2017design} library calls. Both approaches perform far from the machine's peak with average efficiencies of 61\% and 49\% respectively. On the other hand, the orange line exhibits the performance of the vendor-optimized mkl-dnn library that follows strategy (ii) with ad hoc, specialized direct convolution kernels and achieves average efficiency of 81\%, being 1.33$\times$ and 1.64$\times$ faster than the aforementioned generic implementations. However, this performance comes at the cost of complex, specialized kernels that do not generalize to different workloads (e.g.\ RNN/LSTM/MLP) or different architectures (e.g.\ GPUs).

In this work, we introduce a new kernel called \emph{batch-reduce GEMM} and show how the most popular DL workloads and algorithms (RNN/LSTM, CNN and MLP) can be formulated with this new kernel as basic building block. The batch-reduce GEMM kernel essentially multiplies a sequence of input sub-tensor blocks (which form a \emph{batch}) and the partial multiplication results are \emph{reduced} into a single accumulator/output sub-tensor block. Our new kernel is flexible enough to accommodate coarse-grained and fine-grained operations that arise in DL workloads, whereas its semantics lend themselves to various optimizations (e.g.\ load/store optimizations of the result sub-tensor, prefetching of the sub-tensors to be multiplied). Also, since the kernel supports operations at fine granularity, fusion of subsequent operators on the output sub-blocks is inherently efficient. The blue line in Figure~\ref{fig:motivation} shows the performance of the convolution primitive that leverages our new  \emph{batch-reduce GEMM} kernel achieving average efficiency of 83\%, and outperforms even the ad hoc, vendor-optimized kernel.

Having a single kernel as basic building-block is transformative: by implementing and optimizing this single kernel for a given architecture, the development of DL primitives degenerates to mere loop tuning around this kernel. Essentially our approach with a \emph{single} kernel addresses the issue of combinatorial explosion of low-level optimization work that is required for each pair <architecture, DL primitive>. Instead, for each architecture we need to optimize at low-level \emph{only one kernel for all DL primitives}.

Furthermore, having a single, highly efficient building-block enables efficient usage of tensor compiler frameworks. Such frameworks embrace tensors as first class citizens, and provide specific optimization techniques targeting tensor algebra programs. Since DL primitives are inherently tensor algebra programs, there is a large amount of ongoing research that leverages specialized tensor compilers for DL workload development (e.g.\ TVM~\cite{chen2018tvm}, GLOW~\cite{DBLP:journals/corr/abs-1805-00907}, PlaidML~\cite{plaidml}, MLIR~\cite{mlir}). However, compilers struggle to optimize small GEMM-flavored loop nests that arise in tensor programs~\cite{libxsmm}. Contemporary architectures become increasingly complex, and all the micro-architectural idiosyncrasies have to be considered in order to achieve close-to-peak performance. Our kernel is optimized for the nuances of the architecture at hand, and serves tensor compilers a robust building block that can be used during the polyhedral optimization phase of general loop nests~\cite{plaidml,poly}.

To illustrate the viability and generality of our methodology with a single kernel, we develop DL primitives which target training and inference of RNN/LSTM, CNN and MLP workloads in $\sim$3,000 lines of high-level C code. Our primitives outperform vendor-optimized libraries on CPUs. We also provide proof-of-concept design with a tensor compiler framework by showcasing efficient CNN implementation in TVM that leverages our batch-reduce GEMM kernel. Additionally, our methodology provides a pathway for performance portability; we present exemplary, high-performance CNN kernels on integrated GPUs. Last but not least, we integrate our primitives in distributed DL frameworks (Tensorflow~\cite{tensorflow2015} and GxM~\cite{sc18}), and show performance results on two training workloads: Google's Neural Machine Translation (GNMT)~\cite{wu2016google} and ResNet-50 training~\cite{he2016deep}. These results push the envelope of DL training performance on CPU clusters. The main contributions of this paper are:
\begin{itemize}
\item The introduction of the batch-reduce GEMM kernel along with its efficient implementation.
\item The design and implementation of multi-threaded, high performance DL primitives covering RNN/LSTM, CNN and MLP inference and training algorithms with batch-reduce GEMM kernel being the basic building block. We need to optimize at low-level \emph{only this kernel for all DL primitives}.
\item A detailed performance comparison of our DL primitives with state-of-the-art vendor-optimized libraries.
\item Distributed memory results of LSTM and CNN training workloads that leverage our optimized DL kernels and outperform the best in class results on CPU clusters.
\item CNN proof-of-concept results on integrated GPUs and CNN kernels within TVM that leverage the batch-reduce GEMM kernel.
\end{itemize}

\section{The Batch-Reduce GEMM kernel}
\label{sec:br_gemm}
\begin{figure}[t!]
\centering
\includegraphics[width=\columnwidth]{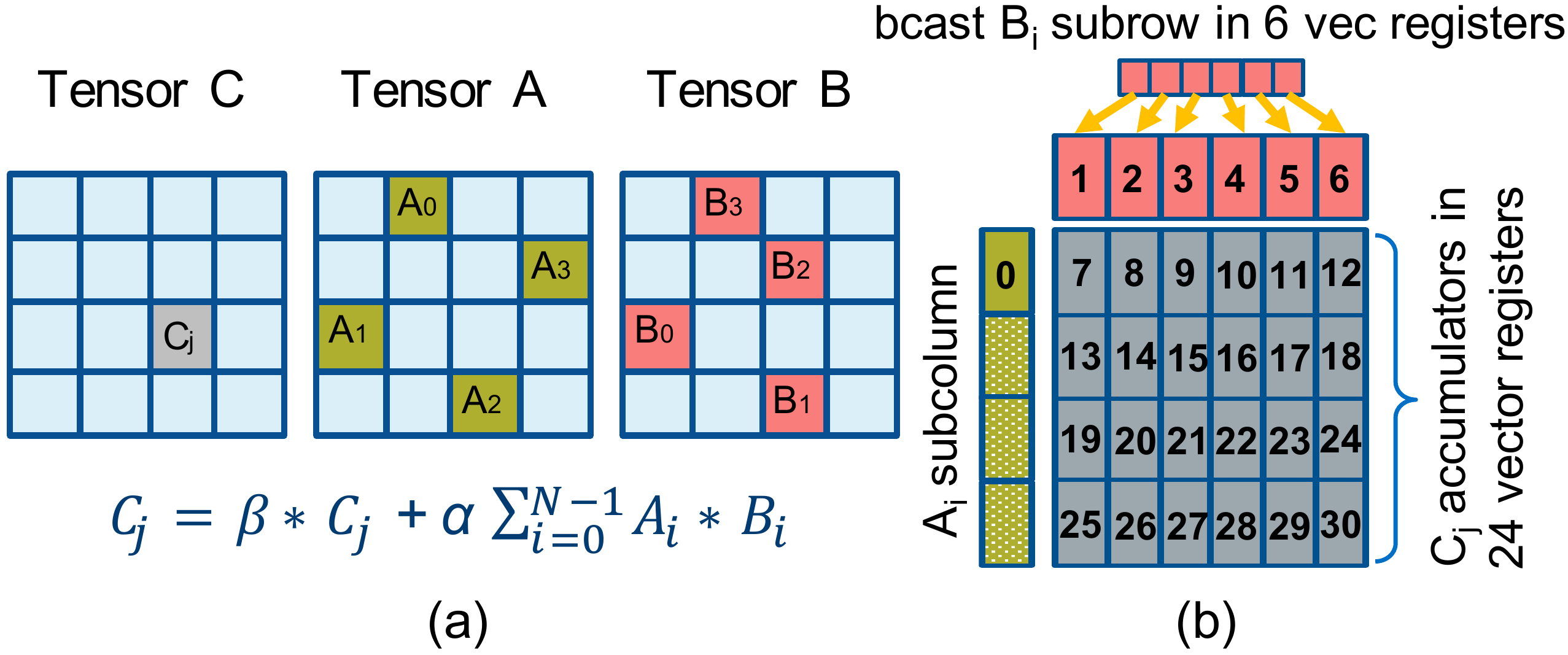}
\caption{(a) The batch-reduce GEMM kernel (b) Outer product small GEMM microkernel}
\label{fig:br_gemm}
\end{figure}

In this section, we describe the design and implementation of the new batch-reduce GEMM kernel which comprises the cornerstone of our deep learning primitives. Figure~\ref{fig:br_gemm} (a) illustrates the functionality of the new kernel which materializes the operation:
\begin{equation*} 
C_j = \beta \cdot C_j + \alpha \sum_{i=0}^{N-1} A_i \cdot B_i
\end{equation*} 
This kernel multiplies the specified blocks $A_i \in {\rm I\!R}^{m\times k}$ and $B_i \in {\rm I\!R}^{k\times n}$ and \emph{reduces} the partial results to a block $C_j\in {\rm I\!R}^{m\times n}$ of a tensor $C$. Tensors $A$ and $B$ can alias and also the blocks $A_i$ and $B_i$ can reside in any position in the input tensors $A$ and $B$. The batch-reduce GEMM kernel takes the following arguments: (i) \emph{two arrays of pointers} to the corresponding blocks $A_i$ and $B_i$ to be multiplied, (ii) a pointer to the output block $C_j$, (iii) the number $N$ of the blocks to be multiplied and (iv) the scaling parameters $\alpha$ and $\beta$. 

Our kernel differs from the recently introduced batched GEMM~\cite{dongarra2017design} and its variation strided-batch-gemm~\cite{stridedbatchgmemm} that materialize:
\begin{equation*} 
C_i = \beta \cdot C_i + \alpha \cdot A_i \cdot B_i
\end{equation*} 
These batched routines are missing the \emph{reduction} functionality and cannot optimize for the output matrix re-use. Also, the strided-batch-gemm kernel accesses the $A_i$ and $B_i$ subblocks based on fixed strides and therefore is more restrictive.

\begin{algorithm}[t]
\begin{algorithmic}[1]
\LState \textbf{Inputs}: $A_i \in {\rm I\!R}^{m\times k},B_i \in {\rm I\!R}^{k\times n}i = 0,...,N\text{-}1$,\ $C_j \in {\rm I\!R}^{m\times n}$ $\alpha, \beta \in {\rm I\!R}$
\LState \textbf{Output}:$\ C_j = \beta \cdot C_j + \alpha \sum_{i=0}^{N-1} A_i \cdot B_i$
\For{$i_n=0 \dots n-1\ \textbf{with\ step\ }\mathbf{n_b}$}
\For{$i_m=0 \dots m-1\ \textbf{with\ step\ }\mathbf{m_b}$}
\State acc\_regs $\leftarrow$  load $m_b \times n_b$ $C_j$$\text{\ subblock}_{i_m,i_n}$ 
\For{$i=0 \dots N-1\ \textbf{with\ step\ } \mathbf{1}$}
\For{$i_k=0 \dots k-1\ \textbf{with\ step\ } \mathbf{1}$}
\LineComment{Outer product GEMM microkernel}
\State acc\_regs $\pluseq$  $A_i\ \text{subcolumn}_{i_m,i_k}\times B_i\ \text{subrow}_{i_k,i_n} $ 
\EndFor
\EndFor
\State  $C_j\ \text{subblock}_{i_m,i_n} \leftarrow$ acc\_regs
\EndFor
\EndFor
\end{algorithmic}
\caption{The batch-reduce GEMM kernel}
\label{alg:br_kernel}
\end{algorithm}

The new batch-reduce GEMM kernel specification naturally lends itself to a handful of optimizations. First, this kernel minimizes the output data movement compared to GEMM or batched GEMM approaches since the specification dictates the use of a single output. Second, the input subblocks that are multiplied can reside in arbitrary locations within tensors, therefore the kernel obviates the need for tensor transformations/copy overheads that are otherwise required in order to obtain long accumulation chains (e.g.\ image to column transformations are required to implement convolutions via large GEMM calls). Such long accumulation chains are essential in order to achieve high performance. Additionally, being able to provide arbitrary sub-tensor blocks as inputs provides ease of integration with blocked/tiled tensor layouts. Last but not least, since the input $A_i$ and $B_i$ subblocks are part of the interface, the implementation can trivially prefetch them in order to hide the latency of data movement.

In order to obtain a high performance implementation of the batch-reduce GEMM kernel we build upon and extend the open source LIBXSMM~\cite{libxsmm} library which leverages JIT techniques and generates small GEMMS achieving close to peak performance. Algorithm~\ref{alg:br_kernel} shows the pseudocode of the batch-reduce GEMM kernel. Lines 1-2 block the computation of the result $C_j$ in $m_b \times n_b$ subblocks. Once such a subblock is loaded into the accumulation registers (line 3), we loop over all pairs $A_i,\ B_i$ (line 4) and we accumulate into the loaded registers the products of the corresponding $m_b\times k$ subblocks of $A_i$ with the relevant $k\times n_b$ subblocks of $B_i$ (lines 5-7). In order to calculate a partial product of an $m_b\times k$ subblock of $A_i$ with a $k\times n_b$ subblock of $B_i$, we follow an outer product formulation. In particular, we multiply an $m_b\times 1$ column of $A_i$ with a $1\times n_b$ row of  $B_i$ (line 7) and we repeat the analogous outer product computation for all $k$ columns/rows of the $A_i$/$B_i$ subblocks (line 5). Figure~\ref{fig:br_gemm}(b) depicts the outer product microkernel that multiplies an $m_b\times 1$ column of $A_i$ with a $1\times n_b$ row of  $B_i$ (in this example $m_b=64$, $n_b=6$). For illustration purposes, we consider that the underlying architecture has 32 vector registers where each one can hold 16 tensor elements. In this example, accumulation registers 7-30 hold the partial $C_j$ result. First, we broadcast the row of $B_i$ into registers 1-6. Then, we load  in register 0 the first 16 elements of the $A_i$ column and via 6 fused-multiply-add instructions (FMAs) with registers 1-6 we update the accumulators 7-12. We repeat the analogous process for the remaining 48 elements of the $A_i$ column and we update all the accumulation registers. We note here that this is just one of the methods that LIBXSMM adopts for the outer product microkernel; LIBXSMM leverages various strategies depending on the architecture at hand (i.e.\ vector length) and the $m_b$, $n_b$ values. Once the $m_b \times n_b$ subblock of $C_j$ is fully computed for all pairs of $A_i$ and $B_i$ matrices, the accumulators are stored in the proper location of $C_j$ (line 8). Finally, we further enhance the microkernel with software prefetches of $A_i$ and $B_i$ elements aiming to mitigate cache miss latency overheads.

\section{Deep Learning Kernels}
\label{sec:dl_algs}
Here we describe the design and implementation of our DL primitives that exploit the batch-reduce GEMM kernel. In particular, we outline how to implement the required algorithms for LSTM~\cite{hochreiter1997long}, CNN~\cite{szegedy2015going} and MLP~\cite{hornik1989multilayer} workloads. We choose performance-optimal data layouts which might differ from classic layout specifications in today's vendor libraries. However, this is fully acceptable as modern DL frameworks anyways change tensor layouts during their graph optimization phase for operator fusion (e.g.\ Tensorflow's Grappler). Therefore, freedom of data layout choice is a fundamental cornerstone to enable high performance through tensor compilers.

We highlight that the subsequent algorithmic descriptions are agnostic of the compute precision.The only prerequisite in order to get an implementation with the desired compute precision is to generate the corresponding batch-reduce GEMM kernel. The results we present in section~\ref{sec:results} are in single precision (FP32), however we already have implementations supporting the int8 and bfloat16 datatypes (via the new Intel VNNI and bfloat16 instructions respectively) which have been shown to sufficiently cover a range of DL training and inference workloads~\cite{vanhoucke2011improving,bfloat16_tf,de2018high} and are supported on up-coming CPU architectures. Also, the same algorithms are applicable for GPUs; in Section~\ref{sec:results} we showcase exemplary results of CNNs on integrated GPUs.
 
\subsection{Long-Short Term Memory (LSTM)}
\label{subsec:lstm}
\begin{figure}[t!]
\centering
\includegraphics[width=\columnwidth]{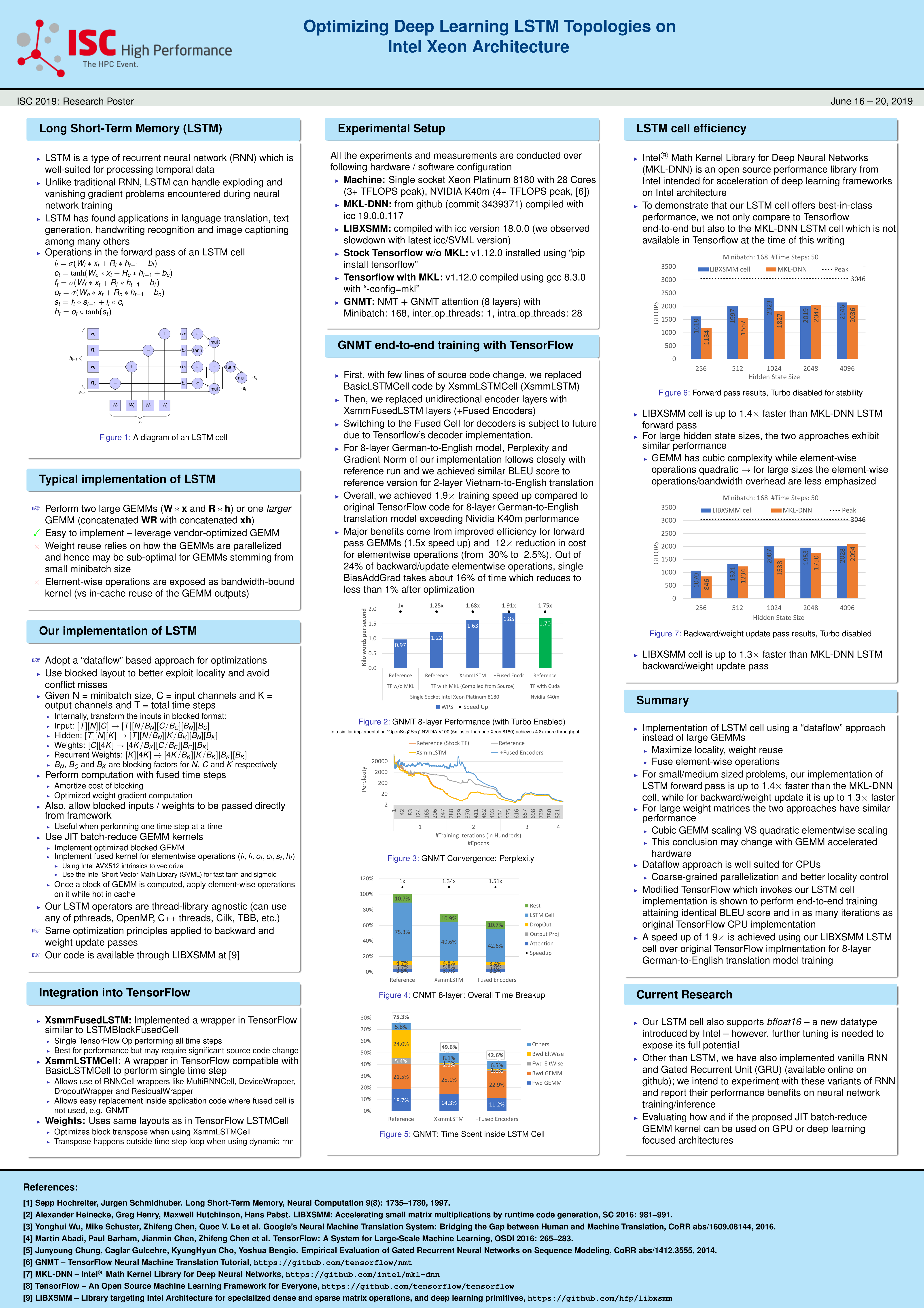}
\caption{Long-Short Term Memory data flow.}
\label{fig:lstm}
\end{figure}

LSTM is a type of RNN which is well-suited for processing temporal data. Unlike traditional RNN, LSTM can handle exploding and vanishing gradient problems encountered during neural network training. LSTM has found applications in language translation, text generation, handwriting recognition and image captioning. In this subsection, we focus on the forward propagation to train an LSTM cell (the forward propagation pass is utilized also for the inference use-case). The backward by data and weight update kernels required for the entire training via the back-propagation algorithm~\cite{lecun1988theoretical} are implemented in an analogous way. 

\subsubsection{LSTM equations and prior art}
\label{subsubsec:naive_lstm}
Given the batch size $N$, the sequence length $T$, the state size $C$ and hidden state size $K$, the inputs of the forward propagation pass in the training process of the LSTM cell are: i) the weight tensors $W_i$, $W_c$, $W_f$, $W_o\in {\rm I\!R}^{K\times C}$, ii) the recurrent weights $R_i$, $R_c$, $R_f$, $R_o\in {\rm I\!R}^{K\times K}$, iii) the input sequence tensor $x \in {\rm I\!R}^{T\times C\times N}$, and iv) the bias tensors $b_i$, $b_c$, $b_f$, $b_o\in {\rm I\!R}^{K}$. These tensors are combined based on the Equations~1-6 and yield the output sequence  $h \in {\rm I\!R}^{T\times K\times N}$ and tensor $s \in {\rm I\!R}^{T\times K\times N}$:
\begin{eqnarray}
i_t &=& \sigma(W_i\cdot x_t + R_i\cdot h_{t-1} + b_i)\\
c_t &=& \textrm{tanh}(W_c\cdot x_t + R_c\cdot h_{t-1} + b_c)\\
f_t &=& \sigma(W_f\cdot x_t + R_f\cdot h_{t-1} + b_f)\\
o_t &=& \sigma(W_o\cdot x_t + R_o\cdot h_{t-1} + b_o)\\
s_t &=& f_t \circ s_{t-1} + i_t \circ c_t\\
h_t &=& o_t \circ \textrm{tanh}(s_t)
\end{eqnarray}
In these equations, observe the \emph{recurrent} relationship between subtensors $i_t, c_t, f_t, o_t$ and $s_t$ of the current time-step $t$ and subtensors $h_{t-1}$, $s_{t-1}$ of the previous time-step $t-1$. Also, $\sigma()$ represents the standard logistic sigmoid function, $\tanh()$ is the hyperbolic tangent function and ``$\circ$" stands for element-wise multiplication of tensors. Figure~\ref{fig:lstm} visualizes the computations and the dependencies involved in the forward propagation pass of the LSTM network.

Typical implementations of the LSTM cell (e.g.\ basic LSTM cell in Tensorflow) stack the $W_i$, $W_c$, $W_f$, $W_o$ matrices into $W\in{\rm I\!R}^{4K\times C}$ and the $R_i$, $R_c$, $R_f$, $R_o$ into $R\in {\rm I\!R}^{4K\times K}$ and then employ two large GEMMS $W\cdot x_t$ and $R\cdot h_{t-1}$ to calculate the relevant partial products in Equations 1-4. Moreover, these two large GEMMs can be further replaced with a single large GEMM call by stacking $W$, $R$ and $x_t$, $h_{t-1}$  and performing: $\begin{bmatrix}  W\ R \end{bmatrix}\cdot \begin{bmatrix}  x_t^T\ h_{t-1}^T\end{bmatrix}^T$. Then, such an implementation applies the element-wise operations (sigmoid/tanh) onto the GEMM results and concludes with the element-wise operations dictated by Equations 5-6. While such an approach is easy to implement by exploiting large vendor-optimized GEMM library calls, the data reuse of the underlying tensors relies on how GEMMs are parallelized and may be suboptimal for GEMM sizes stemming from small batch size $N$. Also, the element-wise operations are exposed as a bandwidth-bound kernel after the GEMM which is typically a compute-bound kernel; the outputs of the large GEMM are not hot in cache (due to limited cache capacity) and as such the involved tensors have to be re-read from memory for the element-wise operations.

\begin{algorithm}[t]
\begin{algorithmic}[1]
\LState \textbf{Inputs}: Weight tensors $W_*[K_b][C_b][b_c][b_k], R_*[K_b][K_b][b_k][b_k]$
\LState Input sequence $x[T][N][C]$, Bias $b_*[K]$, blocking factors $b_k, b_c, b_n$
\LState \textbf{Outputs}: Output sequence $h[T][N][K]$ and $s[T][N][K]$
\State $N_b \leftarrow N/b_n$
\State Based on $thread\_id$ calculate $K_b\_start$, $K_b\_end$, $N_b\_start$ and $N_b\_end$ to assign output work items
\For{$t=0 \dots T-1$}
\For{$ib_k=K_b\_start \dots K_b\_end$}
\For{$ib_n=N_b\_start \dots N_b\_end$}
\LineComment{Compute a block of $i_t = \sigma(W_i\cdot x_t + R_i\cdot h_{t-1} + b_i)$}
\State $i_k \leftarrow ib_k \cdot b_k$\ ,\ $i_n \leftarrow ib_n \cdot b_n$
\State $i[t][i_n][i_k]\leftarrow b_i[i_k]$
\For{$ib_c=0 \dots C_b-1$}
\State $A_{ptrs}[ib_c] = \&W_i[ib_k][ib_c][0][0]$
\State $B_{ptrs}[ib_c] = \&x[t][i_n][ib_c\cdot b_c]$
\EndFor
\State $\mathbf{batchreduce\_gemm}(A_{ptrs}, B_{ptrs}, \&i[t][i_n][i_k], C_b)$ 
\For{$ib_c=0 \dots K_b-1$}
\State $A_{ptrs}[ib_c] = \&R_i[ib_k][ib_c][0][0]$
\State $B_{ptrs}[ib_c] = \&h[t-1][i_n][ib_c\cdot b_k]$
\EndFor
\State $\mathbf{batchreduce\_gemm}(A_{ptrs}, B_{ptrs}, \&i[t][i_n][i_k], K_b)$ 
\State $i[t][i_n][i_k] \leftarrow \sigma(i[t][i_n][i_k])$ 
\LineComment{Ditto for blocks of $c_t, f_t,o_t$ via Equations 2-4}
\State $s[t][i_n][i_k] \leftarrow f[t][i_n][i_k] \circ s[t-1][i_n][i_k] +$
\State \ \ \ \ \ \ \ \ \ \ \ \ \ \ \ \ \ \ \ \ \ \ \ \ \ $i[t][i_n][i_k] \circ c[t][i_n][i_k]$
\State {$h[t][i_n][i_k] \leftarrow o[t][i_n][i_k] \circ \tanh{(s[t][i_n][i_k])}$}
\EndFor
\EndFor
\EndFor
\end{algorithmic}
\caption{Forward propagation pass of LSTM cell}
\label{alg:lstm}
\end{algorithm}

\subsubsection{Optimized LSTM cell via the batch-reduce GEMM kernel}
\label{subsubsec:lstm_opt}
In order to ameliorate the inefficiencies of the large GEMM approach, we follow a \emph{data flow} methodology for our optimized LSTM cell, an approach which has been also explored in previous work~\cite{lstmdataflow}. More specifically, we implement a parallel blocked matrix GEMM in order to achieve load balance, maximize weight matrix reuse and fuse the element-wise operations after partial GEMM blocks are computed and while they are still hot in cache. 

Algorithm~\ref{alg:lstm} shows our data-flow implementation. In particular, the output and the intermediate GEMM results/tensors are divided into logical $b_n\times b_k$ blocks which constitute the work items. Then these work items are assigned onto the available threads (line 2) and subsequently each thread proceeds with its assigned computations. Lines 6-17 indicate how such a $b_n\times b_k$ block of $i_t$ is calculated by a specific thread. First (line 8), the corresponding $i_t$ block is initialized with the according bias tensor values from $b_i$. Then, lines 9-12 employ the batch-reduce GEMM kernel described in Section~\ref{sec:br_gemm} and calculate the contribution $W_i\cdot x_t$ to the current block of $i_t$. More specifically, lines 9-11 prepare the arguments of the batch-reduce GEMM call by calculating the pointers of the required $W_i$ and $x_t$ sub-blocks and storing them in auxiliary arrays $A_{ptrs}$ and $B_{ptrs}$. Then, line 12 calls the batch-reduce GEMM kernel which accumulates the partial products from the $W_i$ and $x_t$ sub-blocks onto the current $i_t$ block. We emphasize here that our batch-reduce GEMM allows small blocking values $b_n$  and $b_k$ to be used since: (a) the small GEMM microkernel runs close to peak even for small dimensions and (b) it avoids the redundant load/stores of the accumulators that arise from the batch-reduce operation and would cripple the overall performance; instead it keeps the accumulation chain in-registers for as long as possible (see Algorithm~\ref{alg:br_kernel}). In an analogous way, lines 13-16 calculate the contribution $R_i\cdot h_{t-1}$ to the current block of $i_t$ as shown in Equation 1. Subsequently, line 17 applies the element-wise operation (sigmoid in this case) onto the just-computed block of $i_t$. Since the block of $i_t$ is hot in cache, the application of the element-wise operation does not incur any data movement from memory.  The same technique is used to calculate the corresponding sub-blocks of $c_t$, $f_t$ and $o_t$ (omitted in Algorithm~\ref{alg:lstm} for simplicity). It is noteworthy that the $c_t$, $f_t$ and $o_t$ computations reuse the same entries of $x_t$ and $h_{t-1}$ from cache since these tensor entries were also used for the computation of $i_t$. Finally, lines 19-21 conclude the computation of the corresponding blocks of the output tensors $h_t$ and $s_t$ based on the element-wise operations dictated by Equations 5-6. After all the work items assigned to the available threads for a given time-step are fully computed, all the threads synchronize and proceed to the next time-step (loop at line 3). Such a synchronization is necessitated because all the output entries $h_t$ of the current time-step are required in the next time-step iteration.

We also note here that the way the work items are processed by the threads affects the data reuse of the weight tensors $W_*$ and $R_*$. In particular, since work items are processed by iterating the ``mini batch" dimension first (loop at line 5), the corresponding slices of the weight tensors $W_*$ and $R_*$ are reused $N_b\_end-N_b\_start-1$ times from cache (potentially from  mid-level cache). Another optimization that is not shown in Algorithm~\ref{alg:lstm} for simplicity is further cache blocking of the batch-reduce loops at lines 9 and 13. In particular, if the weight tensors at hand have large state sizes $C$ and $K$, we block these dimensions in order to fit the corresponding weight tensors slices in cache. In such a case, the algorithm would have yet another loop just after the time-step loop (at line 3) which blocks the batch-reduce loops at lines 9 and 13.

Last but not least, Algorithm~\ref{alg:lstm} carefully chooses the layouts of the corresponding tensors. The weight tensors $W_*$ and $R_*$ are conceptually 2 dimensional tensors, whereas our implementation employs a blocked layout (with $C_b=C/b_c$ and $K_b=K/b_k$) :
\begin{eqnarray*}
W_*[C][K] \rightarrow W_*[K_b][C_b][b_c][b_k],\  R_*[K][K] \rightarrow  R_*[K_b][K_b][b_k][b_k] 
\end{eqnarray*}
Such a blocked layout exposes better locality (i.e.\ the corresponding accesses of weight sub-blocks are \emph{non-strided} with such a layout) and more importantly avoids cumbersome conflict cache misses. Typically the $C$ and $K$ values are large powers of 2 resulting in strided accesses (in the case of the non-blocked format) which are known to cause conflict misses in contemporary associative cache designs. However, our blocked format bypasses this issue by laying out the weight tensors in a format allowing non-strided accesses in the GEMM microkernel. In regard to the activation tensors, we keep the original non-blocked three dimensional format $x[T][N][C]$, $h[T][N][K]$ and $s[T][N][K]$ since strided accesses are barely an issue for the ``B" matrix in the GEMM microkernel (we also confirmed this by experimenting with a blocked format  for the activation tensors). Note that even though our LSTM cell  internally uses a blocked layout for the weight tensors, this does not need to be exposed at the application level; instead, we can transform the weight tensors into the desired blocked layout in the beginning of the algorithm and such a transformation overhead is amortized among the multiple time-steps in the LSTM cell.

Finally, we would like to briefly discuss the importance of a \emph{single}, architecture-specific optimized kernel. All the functionalities in the LSTM cell (forward propagation/backward by data/weight update pass) utilize as building block just our batch-reduce GEMM kernel. The development/parallelization/optimization of the LSTM cell then merely degenerates to tuning/calibrating the surrounding loops around this microkernel -- a process which can be automated to some extent or even implemented in different programming frameworks/tensor compilers like TVM~\cite{chen2018tvm} or PlaidML~\cite{plaidml}.

\label{sec:cnn}
\begin{figure}[t!]
\centering
\includegraphics[width=\columnwidth]{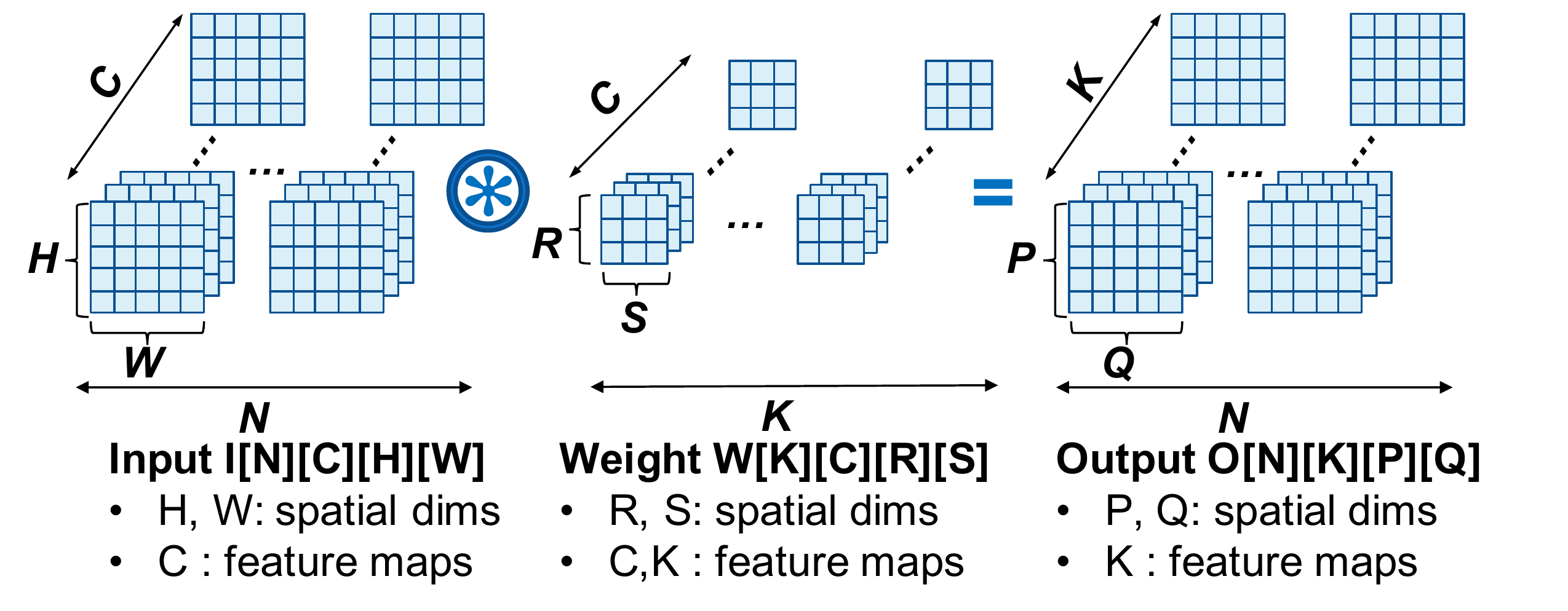}
\caption{Convolution Neural Network (CNN) tensors}
\label{fig:cnn}
\end{figure}
\subsection{Convolution Neural Networks (CNN)}
Convolutional Neural networks (CNN) consist of layers with multiple neurons connected by weights, and have found applications in image recognition, semantic segmentation, autonomous driving and medical imaging. Similar to the LSTM cell, our CNN primitives implement all the kernels required for training via back-propagation. In this section, we describe only the forward propagation kernels which are used as-is for inference. The implementation of backward by data and gradient update kernels follows the same design principles as the forward propagation.  
\subsubsection{Direct convolution loops and prior art}
The values assigned to a neuron are usually called activations. Both activations and weights are represented with multidimensional tensors as illustrated in Figure~\ref{fig:cnn}. The input activation tensors are convoluted with the weight tensors to yield the output activation tensors. The activation tensors conceptually consist of 4 dimensions: the minibatch size $N$, the number of feature maps $C$ and the spatial dimensions $H$ and $W$. We denote the input tensor dimensions with $N$, $C$, $H$ and $W$ while the corresponding output tensor dimensions are $N$, $K$ (output feature maps), $P$ and $Q$ (output spatial dimensions). The weight tensor is conceptually characterized also by 4 dimensions: the feature map dimensions $C$, $K$ and the spatial dimensions $R$ and $S$.

\begin{algorithm}[t!]
\algrenewcommand\algorithmicindent{0.60em}%
\begin{algorithmic}[1]
\State $C_b = C/b_c,\ K_b = K/b_k,\ Q_b = Q/b_q$
\For{$n=0 \dots N-1$}
\For{$k_b=0 \dots K_b-1$}
\For{$c_b=0 \dots C_b-1$}
\For{$oj=0 \dots P-1$}
\For{$oib=0 \dots Q_b-1$}
\State $oi = oib\cdot b_q,\  ii = str\cdot oi,\ ij = str\cdot oj $
\For{$r=0 \dots R-1$}
\For{$s=0 \dots S-1$}
\LineComment{Small GEMM loops}
\For{$k^\prime=0 \dots b_k-1$}
\For{$oi^\prime=0 \dots b_q-1$}
\For{$c^\prime=0 \dots b_c-1$}
\State $oi^{\prime\prime}=oi+oi^\prime$
\State $ij^\prime = ij + r,\ ii^\prime=ii+str\cdot oi^\prime+s$
\State $\scriptsize{O[n][k_b][oj][oi^{\prime\prime}][k^\prime]\pluseq}$
\State $\scriptsize{W[k_b][c_b][r][s][c^\prime][k^\prime]\cdot I[n][c_b][ij^\prime][ii^\prime][c^\prime]}$
\EndFor
\EndFor
\EndFor
\EndFor
\EndFor
\EndFor
\EndFor
\EndFor
\EndFor
\EndFor
\end{algorithmic}
\caption{CNN forward propagation loops}
\label{alg:cnn_basic}
\end{algorithm}

Algorithm~\ref{alg:cnn_basic} shows a basic implementation of the forward propagation loops where the feature map loops (lines 3 and 4) are blocked by factors $b_k$ and $b_c$ respectively and the $Q$ loop (output tensor pixel dimension) is blocked by a factor $b_q$. The input tensor pixels can be also accessed in a strided fashion via a stride $str$. Additionally, the tensors employ a blocked layout format which has been shown to exhibit better locality properties for direct convolutions~\cite{sc18}:
\begin{eqnarray*}
Input\ tensor: & &I[N][C_b][H][W][b_c]\\
Weight\ tensor:& &W[K_b][C_b][R][S][b_c][b_k]\\
Output\ tensor: & &O[N][K_b][P][Q][b_k]
\end{eqnarray*}
By adopting such a blocked layout and given the loop ordering of Algorithm~\ref{alg:cnn_basic}, the three innermost loops (lines 11-17) form a small GEMM of a $b_k\times b_c$ weight sub-tensor with a $b_c\times b_q$ input sub-tensor yielding a $b_k\times b_q$ output subtensor (note that the leading dimension of the input sub-tensor is $str\cdot b_c$). The authors of previous work~\cite{sc18} identified this property; however, they implemented a specialized convolution kernel because:
\begin{itemize}
\item they optimize load/store of the output $O$ in case of $R, S > 1$ and in case the input feature map loop (line 4) is reordered as the innermost loop in order to maximize output reuse.
\item they apply additional pixel blocking when $Q = b_q$ and this value is smaller than the FMA latency of the architecture at hand.
\end{itemize}
In the following subsection, we describe how we address these issues with our new batch-reduce GEMM kernel.

\subsubsection{Optimized convolutions via the batch-reduce GEMM kernel}
\label{subsubsec:cnn_opt}
\begin{algorithm}[t]
\begin{algorithmic}[1]
\State $C_b = C/b_c,\ K_b = K/b_k,\ Q_b = Q/b_q$
\For{$n=0 \dots N-1$}
\For{$k_b=0 \dots K_b-1$}
\For{$c_b=0 \dots C_b-1\ \textbf{with\ step}\ B_c $}
\For{$oj=0 \dots P-1$}
\For{$oib=0 \dots Q_b-1$}
\State $oi = oib\cdot b_q,\  ii = str\cdot oi,\ ij = str\cdot oj,\ i = 0$
\LineComment{Prepare batch-reduce GEMM arguments}
\For{$r=0 \dots R-1$}
\For{$s=0 \dots S-1$}
\For{$c=0 \dots B_c-1$}
\State $A_{ptrs}[i] = \&W[k_b][c_b+c][r][s][0][0]$
\State $B_{ptrs}[i\scriptsize{++}] = \& I[n][c_b+c][ij+r][ii+s][0]$
\EndFor
\EndFor
\EndFor
\State $Out = \&O[n][k_b][oj][oi][0]$ 
\State $\mathbf{batchreduce\_gemm}(A_{ptrs}, B_{ptrs}, Out, R\cdot S\cdot B_c)$ 
\EndFor
\EndFor
\EndFor
\EndFor
\EndFor
\end{algorithmic}
\caption{CNN forward pass via batch-reduce GEMM}
\label{alg:cnn_br_gemm}
\end{algorithm}

The introduction of the batch-reduce GEMM kernel obviates the need for a specialized convolution kernel. More specifically, the batch-reduce GEMM kernel inherently optimizes load/store of the output $O$ in case of $R, S > 1$ and in case the input feature map loop is reordered as the innermost loop. By properly selecting the sub-tensors of weights/inputs to be multiplied and reduced onto an $O$ sub-tensor, the accumulation takes place entirely in registers as described in Section~\ref{sec:br_gemm}. In order to tackle the second issue regarding the case with $Q = b_q$ and $b_q$ being smaller than the FMA latency, we make the following observation: the small GEMM microkernel utilizes $b_q\times (b_k/VLEN)$ accumulator registers where $VLEN$ is the vector length of the architecture at hand. Therefore, if $b_q$ is small then we accordingly increase $b_k$ such that $b_q\times (b_k/VLEN)$ is larger than the FMA latency.

Algorithm~\ref{alg:cnn_br_gemm} shows how to implement the convolution loops using our new batch-reduce GEMM kernel. Note that the input feature map loop (line 4) is blocked by a factor $B_c$ and these $B_c$ iterations are brought into the batch-reduce call in order to further increase the output register reuse. The loops at lines 9-11 prepare the arguments of the batch-reduce GEMM call by calculating the pointers to the weight and input sub-tensors that need to be multiplied and reduced onto a sub-tensor in $O$. In this way, we optimize the output sub-tensor $O$ register reuse: without the batch-reduce kernel we would have to load/store the output registers $(R\times S\times B_c)-1$ additional times. Another optimization involves the case of convolutions with $R=S=1$ and unit stride (i.e.\ $str=1$). In such a case, the input spatial dimensions (loops 5 and 6) are accessed sequentially and as such one can consider that the spatial dimensions are collapsing into a single dimension allowing even more aggressive blocking parameter values $b_q$.

In regard to the parallelization of Algorithm~\ref{alg:cnn_br_gemm}, we observe that the mini-batch dimension (line 2), the output feature map blocks (line 3) and the output pixels blocks (lines 5 and 6) define $N\times K_b\times P\times Q_b$ independent tasks. Typically we opt to divide work first based on the mini-batch dimension since the weight tensors could be reused by multiple threads from shared caches. If we don't have sufficient work just based on the mini-batch size, then we consider all $N\times K_b\times P\times Q_b$ tasks and they are assigned to the available threads in a block fashion. In case our convolution at hand involves large weights, it may be better to assign tasks by starting from the feature map dimension $K_b$. In this way, each thread will touch only a part of the large weight tensor which could be further blocked for a specific cache-level. We implemented all these parallelization strategies and use the most suitable one based on the convolution layer specifications and the available number of threads.

Our backward by data/weight update kernels with batch-reduce GEMM leverage previous work~\cite{sc18}. The authors in~\cite{sc18} show that only slight modifications to the forward kernel are required in order to implement the back-propagation kernels, as they can be mapped through linear index transformations into the forward convolution loop nest (``dual convolutions"). The data reuse optimizations/parallelization tasks then simply translate to tuning the surrounding loops as shown in Algorithm~\ref{alg:cnn_br_gemm}. In Section~\ref{subsec:poc_cnn_results}, we show results of a proof-of-concept design where we develop CNN primitives within a tensor compiler framework via our batch-reduce GEMM kernel. We also show how the same design principles are applicable for integrated GPUs, yielding high performance convolution kernels.

\subsection{Multilayer Perceptron (MLP)}
\label{sec:mlp}

\begin{figure}[t!]
\centering
\includegraphics[width=0.6\columnwidth]{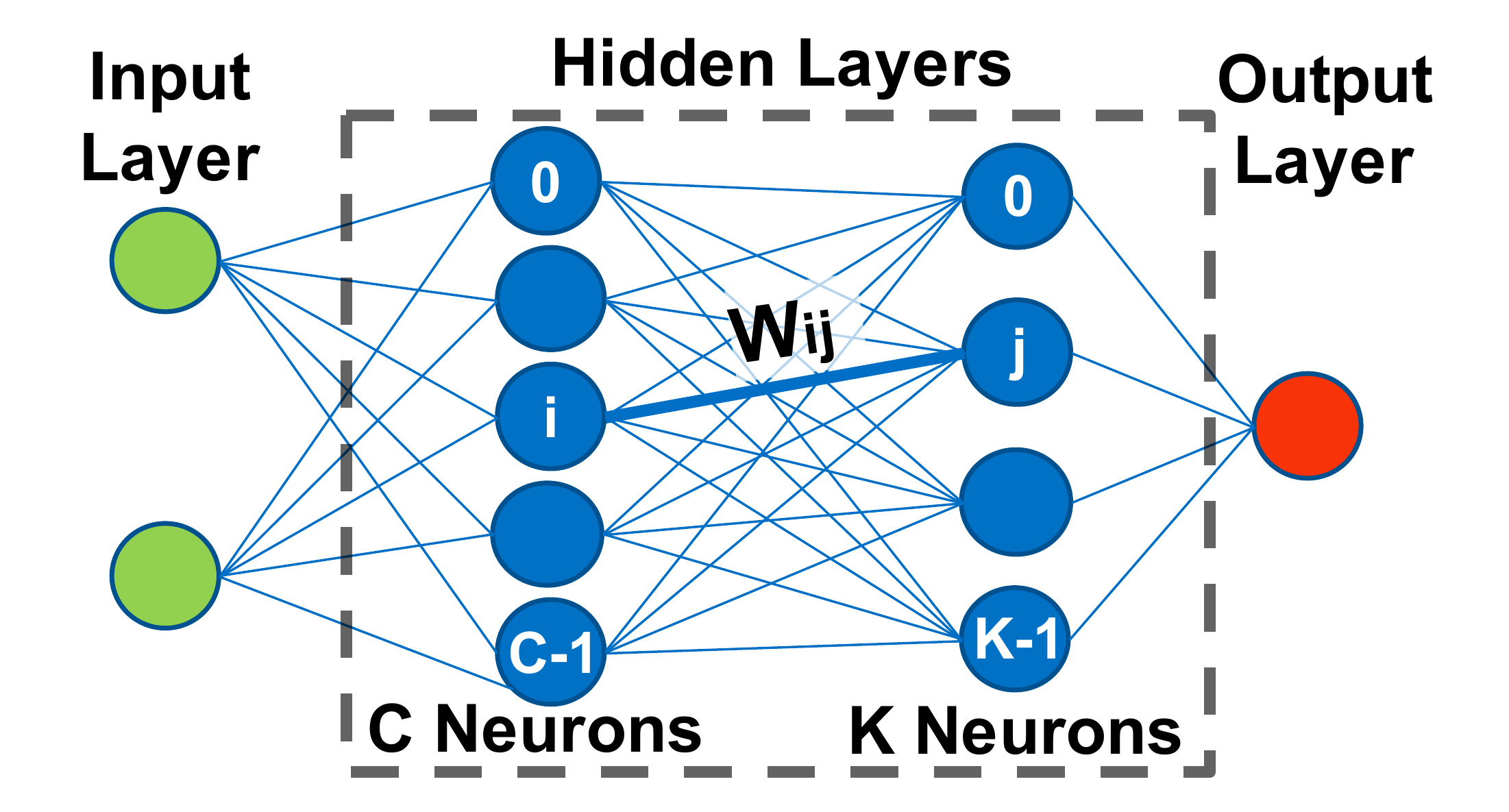}
\caption{A Mulitlayer Perceptron (MLP) topology}
\label{fig:mlp}
\end{figure}
Multilayer perceptrons (MLP) comprise a class of feed-forward artificial neural networks that are widely used for classification tasks, brain modeling, time series prediction, character recognition and data compression. An MLP consists of (at least three) \emph{fully connected} layers of neurons as illustrated in Figure~\ref{fig:mlp}: the topology starts with an input layer, followed by a number of hidden layers which conclude to the output layer. Each neuron in the topology uses a non-linear activation function. For the rest of this section we consider the optimization of the \emph{fully connected} layers since they constitute the cornerstone of MLP. The fully-connected layers also lie in the heart of the modern Transformer~\cite{transformer} and BERT~\cite{bert} workloads. We dive into the details of the forward propagation algorithm of the MLP training process (also used for inference); we also implemented all the required kernels of the back-propagation training in an analogous fashion.

\subsubsection{Fully Connected layers and prior art}
\label{subsubsec:fc_coarse}
The dashed box in Figure~\ref{fig:mlp} illustrates two fully connected layers consisting of $C$ and $K$ neurons respectively. A neuron $i$ from the first layer is connected to a neuron $j$ in the second layer with a weight $W_{ij}$. Mathematically, an input layer $x\in {\rm I\!R}^{C}$ is mapped to an output layer $y\in {\rm I\!R}^{K}$ via the relation $y=W\cdot x$, where $W\in {\rm I\!R}^{K\times C}$ is the weight tensor of the connections between the neurons. During the training process, $N$ multiple inputs ($N$ is the so-called mini-batch size) are grouped together yielding the equation $Y=W\cdot X$ with $W\in {\rm I\!R}^{K\times C}$,  $X\in {\rm I\!R}^{C\times N}$ and $Y\in {\rm I\!R}^{K\times N}$. After the output tensor $Y$ is computed, a non-linear activation function $g()$ is applied on it. Observe that by increasing the mini-batch $N$, we fundamentally increase the weight tensor reuse. Typical implementations of Fully Connected layers leverage a large GEMM call and they apply the activation functions onto the GEMM outputs. Even though such an approach is straightforward to implement, its performance can be underwhelming for three reasons: i) typical high-performance GEMM library calls internally perform packing of sub-matrices to ameliorate TLB misses and cache conflict misses~\cite{goto2008anatomy}, ii) the multi-threaded implementation of GEMM with shapes arising from small mini-batch values $N$ may not fully exploit the available data reuse, and iii) in case of large matrices that do not fit in cache, the activation function application is exposed as a bandwidth-bound kernel which decays the overall performance. In the next subsection, we describe how our implementation of Fully Connected layers via the batch-reduce GEMM kernel addresses all these issues.

\begin{algorithm}[t]
\begin{algorithmic}[1]
\LState \textbf{Inputs}: Weight $W[K_b][C_b][b_c][b_k]$, Input $X[N_b][C_b][b_n][b_c]$
\LState \textbf{Outputs}: Output $Y[N_b][K_b][b_n][b_k]$
\State Based on $thread\_id$ calculate $K_b\_start$, $K_b\_end$, $N_b\_start$ and $N_b\_end$ to assign output work items
\For{$ib_n=N_b\_start \dots N_b\_end$}
\For{$ib_k=K_b\_start \dots K_b\_end$}
\LineComment{Prepare batch-reduce GEMM arguments}
\For{$ib_c=0 \dots C_b-1$}
\State $A_{ptrs}[ib_c] = \&W[ib_k][ib_c][0][0]$
\State $B_{ptrs}[ib_c] = \&X[ib_n][ib_c][0][0]$
\EndFor
\State $Out =  \&Y[ib_n][ib_k][0][0]$
\State $\mathbf{batchreduce\_gemm}(A_{ptrs}, B_{ptrs}, Out, C_b)$ 
\State $Y[ib_n][ib_k][0][0] \leftarrow g(Y[ib_n][ib_k][0][0])$
\EndFor
\EndFor
\end{algorithmic}
\caption{Forward pass of Fully Connected Layer}
\label{alg:mlp}
\end{algorithm}

\subsubsection{Fully Connected layers via the batch-reduce GEMM kernel}
\label{subsubsec:fc_opt}
Algorithm~\ref{alg:mlp} shows the implementation of the forward propagation in the training process of fully connected layers. First, we highlight the blocked tensor layout; all the 2 dimensional tensors are transformed into 4 dimensional ones by blocking the mini-batch dimension $N$ with a factor $b_n$ and the tensor dimensions $C$ and $K$ with blocking factors $b_c$ and $b_k$ respectively. Such a blocked layout addresses issue (i) mentioned in the previous subsection by exposing better locality and avoiding large, strided sub-tensor accesses which are known to cause TLB misses and cache conflict misses in case the leading dimensions are large powers of 2.

Our algorithm first assigns the output sub-tensor blocks to the available threads (line 1) and every thread then for each assigned output $Y$ block calculates the addresses of the $W$ and $X$ sub-tensor blocks that need to be multiplied and reduced onto the current output $Y$ block (lines 5-7). Note that our JIT-ed kernel allows small values of blocking values $b_n$ to be used, and as such we can extract parallelism from the mini-batch dimension even for small values of $N$. By following the loop ordering of Algorithm~\ref{alg:mlp}, a weight sub-tensor is reused by each thread $N_b\_end-N_b\_start-1$ times, potentially from some level of cache. Also, multiple threads are able to read weights from shared caches when the assigned $Y$ blocks correspond to the same subspace of the $K$ dimension. Finally, in case a weight sub-tensor does not fit in the targeted/desired level of cache, we can further block loops at lines 3 and 5. These cache blocking techniques in combination with the flexible blocking factors $b_n$, $b_c$ and $b_k$ which yield high performance micro-kernels, address the data reuse issue (ii) mentioned in the previous subsection.

Finally, once the arguments of the batch-reduce GEMM have been calculated, we perform the batch-reduce GEMM call (line 9) and while the output sub-tensor block $Y$ is still hot in cache we apply on it the relevant activation function (line 10). In this way, we ensure that the application of the activation function takes place when the data are still hot in cache and it does not incur any additional data movement from memory, addressing issue (iii) from the previous subsection. Once again, the development of the Fully Connected primitive follows the same recipe as the LSTM and CNN primitives. Therefore, the loops surrounding the batch-reduce GEMM kernel can be automatically optimized with a tensor compiler/infrastructure.

\begin{figure*}[t!]
\centering
\includegraphics[width=2.0\columnwidth]{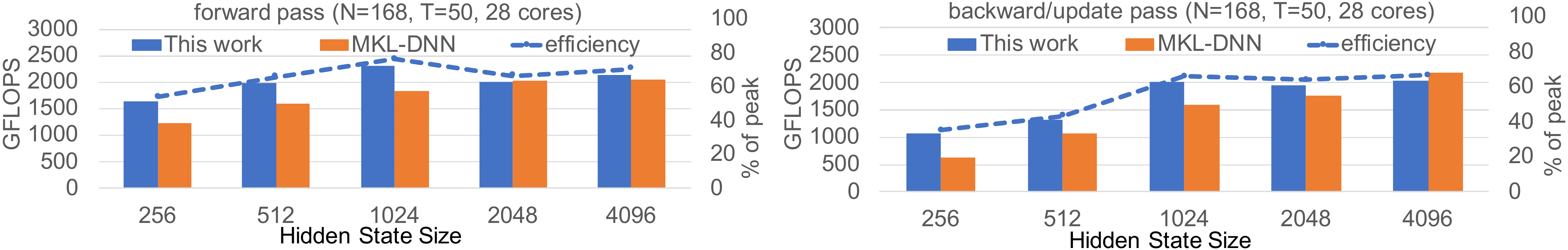}
\caption{Performance of LSTM cell: (Left) Forward propagation and (Right) backward by data and weight update pass.}
\label{fig:lstm_cell}
\end{figure*}

\section{Performance results}
\label{sec:results}
In subsection~\ref{subsec:dl_perf}, we evaluate the performance of our DL kernels. Then, in subsection~\ref{subsec:distr_training}, we present distributed memory results on two state of the art workloads, namely Google's Neural Machine Translation (GNMT) which corresponds to LSTM training and ResNet-50 which is representative of CNN training. 
Finally, in subsection~\ref{subsec:poc_cnn_results}, we show a couple of proof-of-concept results that highlight the generalizability of our approach. More specifically, we show CNN kernel results on integrated GPUs and conclude with CNN kernel results that are generated by TVM, both leveraging batch-reduce GEMM kernel as their basic building block.

\subsection{Performance evaluation of our DL kernels}
\label{subsec:dl_perf}
Since we use a JIT-ing methodology for the batch-reduce GEMM kernel, we can virtually run on every platform supporting SSE, AVX, AVX2 and AVX-512 instructions. All the experiments presented in this subsection are conducted on a Skylake-SP (SKX) 8180 processor with 28 cores, 96 GB\,DDR4 2666 main memory at 2.3\,GHz (AVX512) Turbo at 205W TDP. The stream triad of a single socket is 105\,GB/s. 

For the experiments we used all 28 cores with turbo disabled (i.e.\ AVX-512 base frequency at 1.7\,GHz) in order to get stable measurements. With such a setup, the peak of the machine is $\sim$3,050 GFLOPS (single precision). All the experiments were performed 400 times and we report the average timing; due to careful configuration of our platform (i.e.\ tick-less Linux kernel, core pinning, turbo disabled) the run-to-run variation is within 3\%. For our work we used the Intel compilers (version 18.0.0). For performance comparisons, we used the latest version of MKL-DNN (version 0.9).

\subsubsection{Performance evaluation of LSTM cell}
\begin{table}[!tb]
\fontsize{8}{8}\selectfont
\centering
\begin{tabular}{|c|c|c|c|c|c|c|}
    \hline
        &
      \multicolumn{2}{c|}{{\textbf{batch-reduce GEMM
            }}} & 
      \multicolumn{1}{c|}{{\textbf{Elementwise
            }}} & 
      \multicolumn{1}{c|}{{\textbf{Tensor
            }}} \\
    {\textbf{LSTM pass}} & \textbf{\% of total} & \textbf{GFLOPS} & \textbf{operations} & \textbf{reformatting }\\
    \hline
    \hline
    \textbf{fwd} &  93.3\%  &  2550  &   5.3\% &  1.4\% \\
    \hline
    \textbf{bwd \& upd}  &  91.2\%  &  2350  & 5.3\% & 3.5\% \\
    \hline
  \end{tabular}
\caption{\label{tab:lstm_breakdown}Breakdown of LSTM cell performance ($\mathbf{C}$=$\mathbf{K}$=1024).}
\end{table}

Figure~\ref{fig:lstm_cell}~(Left) illustrates the performance of the forward propagation algorithm in the LSTM cell that is described in subsection~\ref{subsec:lstm}. In this experiment, we fix $N=168$ (mini-batch), $T=50$ (sequence length), and we vary the hidden state size $K$ which is equal to the input state size (i.e.\ $C=K$). The blue bars represent the performance in GFLOPS  (see Left y-axis) of our kernels. We observe that even in the case of small $C$ and $K$, our LSTM cell runs at $\sim$60\% of peak (see Right y-axis), whereas for larger weight tensors the activation-tensor reuse is larger and consequently the kernels run at $\sim$70\% of peak. In Table~\ref{tab:lstm_breakdown} (row labeled ``fwd'') we provide more details regarding how the time is spent within the LSTM cell for the case with $C=K=1024$. During the forward pass, 93.3\% of the time is spent in the batch-reduce GEMM kernel which runs at 2550 GFLOPS or equivalently at 84\% of peak. Then, 5.3\% of the execution time is spent for the elementwise operations described in subsection~\ref{subsubsec:lstm_opt}. The rest 1.4\% is spent in reformatting the weight tensors to take advantage of the blocked format (see subsection~\ref{subsubsec:lstm_opt}).

Figure~\ref{fig:lstm_cell}~(Right) exhibits the performance of the remaining two passes in the LSTM training process, namely backward propagation and weight update pass (henceforth called ``bwd'' and ``upd'' respectively). The performance follows the same trend as the forward propagation, i.e.\ with larger  weight tensors, the overall efficiency is closer to peak due to more re-use of the activation tensors. Notably, the overall efficiency is diminished compared to the forward propagation; by inspecting the time breakdown at Table~\ref{tab:lstm_breakdown} (row labeled ``bwd \& upd'') we observe that larger fraction of the overall time is spent in tensor reformatting. This is expected because bwd and upd passes require algorithmically additional weight and activation tensor transposes~\cite{lecun1988theoretical}. Also, the batch-reduce GEMM runs at 2350 GFLOPS or equivalently at 77\% of peak, which is a bit lower than the efficiency of the one achieved in the forward pass. This is a result of different tensor shapes in  the ``upd'' pass, where the reduction dimension of GEMM becomes the mini-batch dimension and typically it is smaller than $C$/$K$ which constitute the GEMM reduction dimensions in forward pass.

In Figure~\ref{fig:lstm_cell}, we also provide performance comparison of our LSTM cell with the vendor-optimized LSTM cell within MKL-DNN (orange bars). For small to medium problem sizes, our LSTM cell is faster than the MKL-DNN cell in the range of $1.2$-$1.3\times$ for forward propagation and $1.1$-$1.7\times$ for the bwd/upd pass. This is a result of the adopted ``data-flow'' approach described in subsection~\ref{subsec:lstm} that leverages our batch-reduce GEMM kernel: the elementwise operations are naturally fused within the GEMM operations which run at high efficiency. For larger problem sizes, the overall cost of the GEMM operation dominates the entire computation and as such the elementwise operations are negligible. This is expected since the GEMM computation cost scales cubically compared to the quadratic scaling of the elmentwise operations. Therefore, for large problem sizes a coarse grained approach like the one described in subsection~\ref{subsubsec:naive_lstm} yields good performance. It is worth mentioning that in the following subsection~\ref{subsec:distr_training} where we present distributed memory GNMT training results, the involved LSTM corresponds to the case with $C$=$K$=1024 in Figure~\ref{fig:lstm_cell}, where our code is $1.26\times$ faster than MKL-DNN for all training passes (for $N$=168).

\subsubsection{Performance evaluation of CNN kernels}
\begin{figure*}[t!]
\centering
\includegraphics[width=2.0\columnwidth]{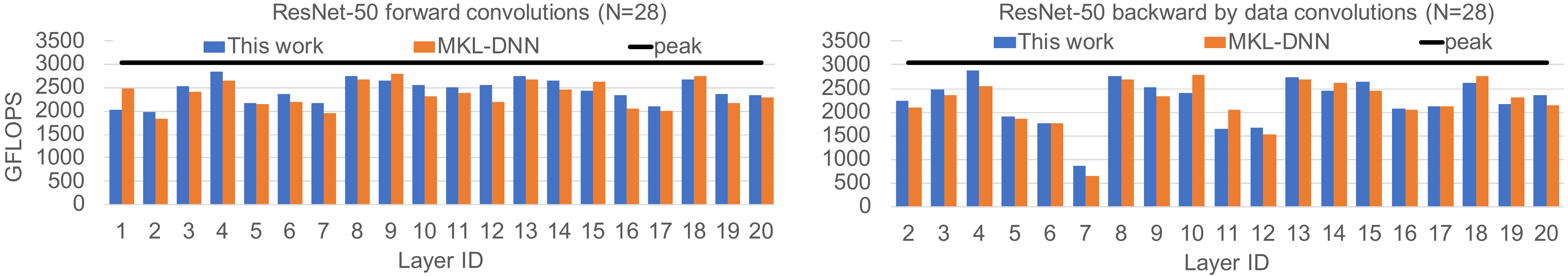}
\caption{Performance of ResNet-50 convolutions: (Left) Forward propagation and (Right) backward by data pass.}
\label{fig:cnn_fwdbwd}
\end{figure*}

\begin{figure}[t!]
\centering
\includegraphics[width=\columnwidth]{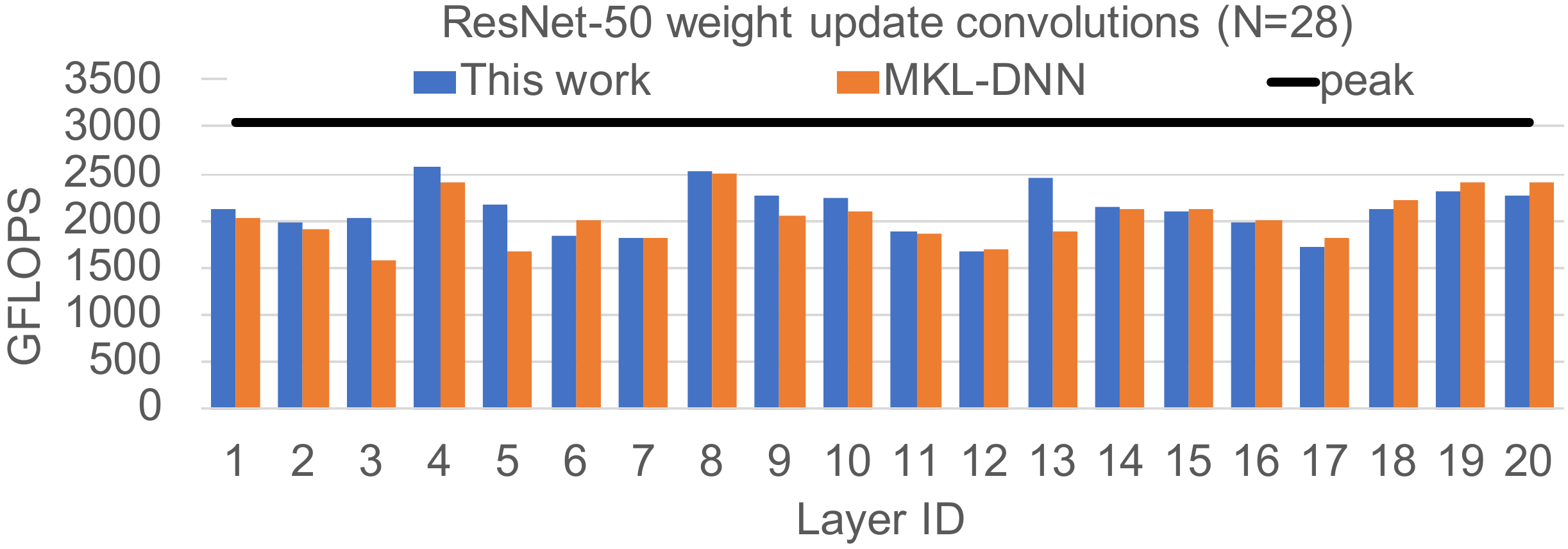}
\caption{Performance of ResNet-50 weight update pass.}
\label{fig:cnn_upd}
\end{figure}

\begin{table}
\fontsize{6}{5}\selectfont
 \begin{tabularx}{\columnwidth}{|X| l | l |X|X|X|X| X || X| l | l |X|X|X|X| X |}
 \hline
ID	&	C	&	K	&	H	&	W	&	R	&	S	&	str	&	ID	&	C	&	K	&	H	&	W	&	R	&	S	&	str	\\
 \hline
 \hline
1	&	3	&	64	&  \fontsize{6}{4}\selectfont 224	&\fontsize{6}{4}\selectfont	224	&	7	&	7	&	2	&	11	&	512	&	1024	&	28	&	28	&	1	&	1	&	2	\\
  \hline
2	&	64	&	256	&	56	&	56	&	1	&	1	&	1	&	12	&	512	&	256	&	28	&	28	&	1	&	1	&	2	\\
 \hline
3	&	64	&	64	&	56	&	56	&	1	&	1	&	1	&	13	&	256	&	256	&	14	&	14	&	3	&	3	&	1	\\
 \hline
4	&	64	&	64	&	56	&	56	&	3	&	3	&	1	&	14	&	256	&	1024	&	14	&	14	&	1	&	1	&	1	\\
 \hline
5	&	256	&	64	&	56	&	56	&	1	&	1	&	1	&	15	&	1024	&	256	&	14	&	14	&	1	&	1	&	1	\\
 \hline
6	&	256	&	512	&	56	&	56	&	1	&	1	&	2	&	16	&	1024	&	2048	&	14	&	14	&	1	&	1	&	2	\\
 \hline
7	&	256	&	128	&	56	&	56	&	1	&	1	&	2	&	17	&	1024	&	512	&	14	&	14	&	1	&	1	&	2	\\
 \hline
8	&	128	&	128	&	28	&	28	&	3	&	3	&	1	&	18	&	512	&	512	&	7	&	7	&	3	&	3	&	1	\\
 \hline
9	&	128	&	512	&	28	&	28	&	1	&	1	&	1	&	19	&	512	&	2048	&	7	&	7	&	1	&	1	&	1	\\
 \hline
10	&	512	&	128	&	28	&	28	&	1	&	1	&	1	&	20	&	2048	&	512	&	7	&	7	&	1	&	1	&	1	\\
 \hline
\end{tabularx}
\caption{ResNet-50 layers specifications}
\label{tab:resnet_layers}
\end{table}

We conducted experiments with the ResNet-50 topology which yields state of the art results in image recognition tasks~\cite{he2016deep}. The convolution layers within the ResNet-50 topology cover a wide variety of parameters/configurations (e.g.\ filter dimensionality and sizes, input sizes, strided convolutions) and can be seen at Table~\ref{tab:resnet_layers}. In this Table we assign to each convolution layer an ID that is used as identifier in the performance plots. Also, for the remaining of this paper we will use the term \emph{weighted efficiency} when presenting ResNet-50 results; each layer $i$ of Table~\ref{tab:resnet_layers} requires $F_i$ flops, takes $t_i$ seconds to be computed, and is represented $n_i$ times in the entire topology (which has 53 layers in total). The weighted efficiency of the entire topology is given by: $(\sum_{i=0}^{52}n_i\cdot F_i)/(\sum_{i=0}^{52}n_i\cdot t_i)$.

Figures~\ref{fig:cnn_fwdbwd} and ~\ref{fig:cnn_upd} show the performance of the ResNet-50 convolutions with mini-batch size $N=28$. The blue bars in Figure~\ref{fig:cnn_fwdbwd} (Left) represent the achieved performance (in GFLOPS) of the forward (FWD) propagation algorithm described in subsection~\ref{subsubsec:cnn_opt} that leverages the batch-reduce GEMM kernel. The weighted efficiency of the FWD convolutions within the ResNet-50 topology is 83\% of peak. More specifically, the convolutions with large spatial filters (e.g.\ $R$=$S$=$3$ in convolutions with IDs 4, 8, 13, 18) run at $\sim$90\% of peak since they inherently have more input and output tensor reuse than the convolutions with $R$=$S$=$1$ which run at $\sim$80\% of peak. Notably, layer with ID 2 runs at 65\% of peak since it has large output spatial and feature map dimensions and as such its performance is bound by the write bandwidth of our system. By comparing the performance of our kernels to MKL-DNN (orange bars) we observe similar trends. The MKL-DNN library exhibits weighted efficiency of 81\% of peak for FWD convolutions and as such it is 2.5\% slower than our work. This result highlights the effectiveness of our approach with a single kernel as basic building block: our convolutions consist of just $\sim$1500 lines of code (for all training passes) whereas the convolution portion of the MKL-DNN library is $\sim$36000 lines of code since it leverages ad hoc, specialized kernels (e.g.\ ad hoc optimization of the direct convolution loops, different approaches/code generation for various $R$ and $S$ values).

Figures~\ref{fig:cnn_fwdbwd} (Right) and ~\ref{fig:cnn_upd} exhibit the performance of the convolution kernels in the remaining training passes, namely backward by data (BWD) and weight update (UPD). Our kernels have weighted efficiency 80\% and 73.6\% for the BWD pass and the UPD pass respectively. Similarly to the convolutions in the FWD pass, the layers with large spatial weight dimensions show better performance than the ones with $1\times 1$ spatial dimensions since the former have better input and output tensor reuse properties. Also, we note that the efficiency of the UPD kernels is $\sim$10\% lower than the efficiency of FWD/BWD kernels. This is a consequence of the weight tensor reduction which is required in the weight update algorithm in order to maximize the input and output tensor data movement~\cite{sc18}. For comparison, the MKL-DNN BWD and UPD kernels illustrate weighted efficiencies of 78.9\% and 68.9\% respectively, and are 1\% and 7\% slower than our kernels. In subsection~\ref{subsec:distr_training} we integrate our kernels in the GxM distributed framework and improve the best in class performance of ResNet-50 training on CPUs.

\subsubsection{Performance evaluation of Fully Connected Layers}
\begin{figure*}[t!]
\centering
\includegraphics[width=2.0\columnwidth]{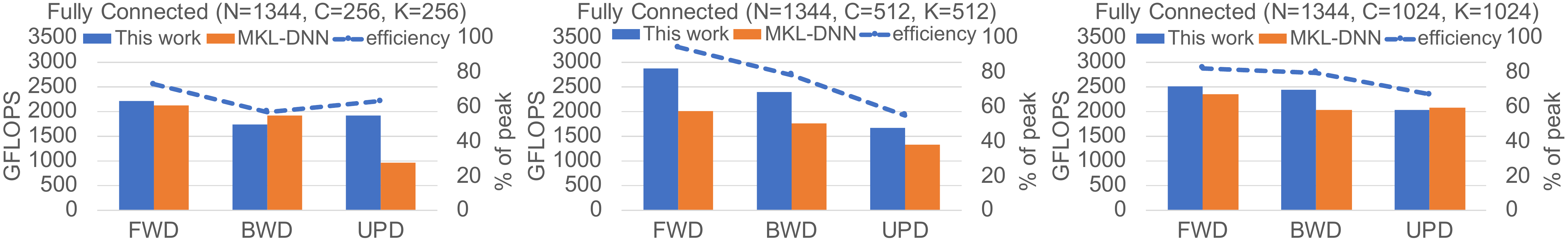}
\caption{Performance of Fully Connected Layers. Bars correspond to Left y-axis / efficiency corresponds to Right y-axis.}
\label{fig:mlp_perf}
\end{figure*}
Figure~\ref{fig:mlp_perf} shows the performance of the Fully Connected layers which are the cornerstone of the MLP workload. In these experiments we fix the mini-batch size $N$=1344 and we vary the dimensions of the weight tensors. For each configuration, we show results for the forward propagation (FWD), backward by data pass (BWD) and weight update pass (UPD). We observe that our approach (blue bars) with the fine-grained batch-reduce GEMM kernel shows for the smaller configuration ($C$=$K$=256) efficiencies in the range 57\%-73\%, for the medium weight sizes ($C$=$K$=512) 55\%-94\% and for the larger configuration ($C$=$K$=1024) the efficiencies are 67\%-82\%. In all configurations, we observe that the BWD kernels's performance deteriorates compared to the equivalent FWD kernels. This is the case because the BWD kernels require a weight transpose~\cite{lecun1988theoretical}. The overhead of this weight transpose is more emphasized in the cases with small $C$/$K$ values while for the cases with large $C$/$K$ values the cost of the GEMM kernel dominates the overall runtime and as such the transpose cost in negligible (see case with $C$=$K$=1024). In regard to the UPD kernels, we also observe that for smaller weight tensors the performance is lower than the corresponding FWD kernels. This is due to the limited parallelism that is available in such cases. More precisely, the FWD pass employs parallelism in the $N$/$K$ dimensions (see Algorithm~\ref{alg:mlp}), the BWD pass in the $N$/$C$ dimensions while the UPD pass in the $C$/$K$ dimensions. Consequently it is more challenging to extract sufficient parallelism within the configurations with small $C$/$K$ values.

Moreover, Figure~\ref{fig:mlp_perf} shows the performance of the Fully Connected layers within the MKL-DNN library (orange bars). These kernels use the coarse-grained approach (i.e.\ a single large GEMM call) as described in subsection~\ref{subsubsec:fc_coarse}. Considering the average efficiencies of all MKL-DNN kernels (FWD, BWD and UPD), the smallest configuration achieves 55\% of peak, the medium configuration runs at 56\% of peak and the largest test case attains 70\% of peak. In contrast, our approach with the batch-reduce GEMM kernel achieves 64\%, 76\%, and 76\% of peak respectively and is $1.16\times$, $1.36\times$, and $1.09\times$ faster than the corresponding coarse-grained approach.

\subsection{Distributed memory training results}
\label{subsec:distr_training}

\begin{figure*}[t!]
\centering
\includegraphics[width=2.0\columnwidth]{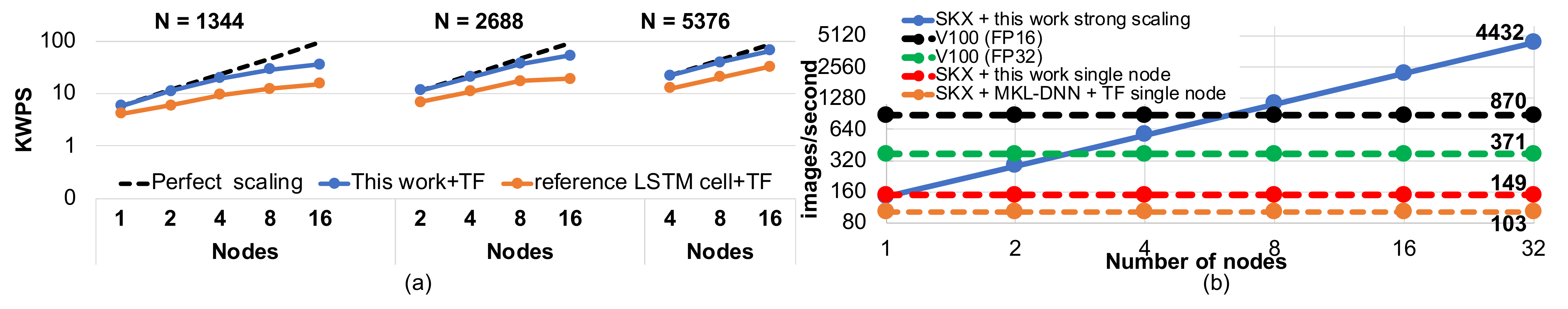}
\caption{Distributed memory training results: (a) 4-Layer GNMT model (LSTM kernels), (b) ResNet-50 model (CNN kernels) }
\label{fig:strong_all}
\end{figure*}

Our experimental platform is a 32-node cluster (Intel Omnipath interconnect), each node having two Skylake-SP (SKX) 8180 processors. For these runs, we enable the turbo mode on the processors (i.e.\ clock frequency up to 2.3\,GHz).

\subsubsection{Distributed memory GNMT training results}
We conducted our experiments with the 4-layer GNMT~\cite{wu2016google} model. The framework of our choice is Tensorflow (TF)~\cite{tensorflow2015}, where we replaced the Tensorflow LSTM cell implementation with our optimized LSTM cell that leverages the batch-reduce GEMM kernel.
Then, we utilized Uber's Horovod library~\cite{horovod} to enable efficient multi-node runs. In order to accelerate the communication performance of Horovod, we replaced its default MPI communication backend with Intel's MLSL library~\cite{mlsl} which optimizes communication primitives arising in deep learning. Moreover, we extend the partitioning logic of the inputs by grouping sequences with similar length together in order to achieve load balance; such a technique yields up to 1.5$\times$ speedup compared to classic input partitioning. For all the experiments in this section, we use 1 MPI rank per CPU socket. As a baseline for comparisons, we used the default LSTM cell for CPUs within TF which we configured to use the MKL library to materialize efficiently the large GEMM calls. In order to assess the benefits of our new kernels, we incorporated our LSTM cell and we further modified the TF code to support fused time-step operations as they are described in Algorithm~\ref{alg:lstm}. We verified correctness of the code changes by achieving state of the art BLEU score of 22.7 after 3 epochs with the German to English WMT16 dataset~\cite{wmt16}.
 
Figure~\ref{fig:strong_all}~(a) illustrates the strong scaling of the distributed memory training with three different batch sizes; the usage of such large batch sizes (up to $\sim$ 5K) is enabled by the LEGW~\cite{you2019large} approach. The y-axis represents the achieved training performance in Kilo Words per Second (KWPS) while the x-axis shows the number of nodes. Both axes are in logarithmic scale. For the smaller batch size ($N$=1344), our approach (solid blue line) scales from 1 to 4 nodes with 84\% strong scaling efficiency, and when we keep scaling from 4 to 16 nodes the parallel efficiency further drops down to 38\%. The main reason for this efficiency drop involves the small mini-batch per socket as we strong scale (we use pure data parallelism to scale out). As a result, we get reduced efficiency within the LSTM cell computation. For example, with $N$=1344 and at the concurrency of 1 node (2 MPI ranks), the mini-batch per socket is 672 whereas at 16 nodes (32 MPI ranks), the mini-batch per socket is 42. Nevertheless, we are able to increase the performance all the way up to 16 nodes even for such a small batch size and we achieve 35.8 KWPS. For comparison, the reference LSTM cell + TF approach (orange line) achieves 15.36 KWPS, thus the approach with our kernels is 2.33$\times$ faster. As we increase the global batch size, the strong scaling efficiency is better because the local computation does not suffer from very small mini-batch. For example, with batch size $N$=2688 our strong scaling efficiency at 16 nodes is 58\% achieving 52.5 KWPS, and is 2.77$\times$ faster than the reference LSTM cell that achieves 18.9 KWPS. With the largest batch size ($N$=5376) the strong scaling efficiency at 16 nodes is 75.2\% achieving 65.9 KWPS. For the same setup, the reference LSTM cell achieves 32.32 KWPS, thus our approach is 2.04$\times$ faster. For the last two experimental setups, we start from 2 and 4 nodes respectively since those batch sizes are too large for the available memory of a single node. To the best of our knowledge, these achieved GNMT training results are the best in class on CPU platforms. For completeness, we mention here the performance of contemporary Nvidia V100 GPU systems on the 4-layer GNMT model in Tensorflow (FP32 precision): The achieved performance is 12.7 and 83.3 KWPS on 1 and 8 GPUs respectively~\cite{v100gnmt}. The FP32 peak performance of V100 GPU is 15.7 TF/s which is $\sim$2$\times$ larger than a single CPU node, and also the available bandwidth is 900 GB/s which is 3.6$\times$ larger than the bandwidth of our node. The scaling from 1 to 8 GPUs shows similar scaling efficiency to our distributed training results.

\subsubsection{Distributed memory ResNet-50 training results}
For ResNet-50 training, we integrated our new CNN kernels into the GxM framework which has been shown to scale efficiently on clusters of CPUs via the MLSL library~\cite{sc18,mlsl}. We verified correctness of our experiments by converging to the same state of the art accuracies, e.g.\ 75.7\% Top-1 accuracy. For Nvidia V100 performance, we use numbers from Nvidia~\cite{v100_fp16_resnet} and previous work~\cite{xu2018deep}.
 
Figure~\ref{fig:strong_all}~(b) summarizes the obtained performance. For the single node CPU measurements, we use 28 cores per socket and the mini-batch is 56 (dual socket nodes). Our approach (red flat line) achieves 149 images/sec and is 1.45$\times$ faster than the configuration with MKL-DNN and Tensorflow (orange flat line) which attains 103 images/sec. If we increase the mini-batch to 224, MKL-DNN improves its efficiency and is able to obtain 129 images/sec. For completeness, we mention the achieved performance of one V100 GPU which is 371 images/sec in single precision (green flat line) and 870 images/sec in mixed FP16/FP32 precision (black flat line) -- the latter approach uses the available tensor cores~\cite{markidis2018nvidia}.

In order to scale out with GxM, we dedicate 2 cores per node for communication via MLSL primitives and we use 54 cores for computations. In Figure~\ref{fig:strong_all}~(b), we illustrate with solid blue line the scaling of GxM with our new CNN kernels (both axes are in logarithmic scale). We scale up to 32 nodes with 95.3\% parallel efficiency, achieving at the concurrency of 32 nodes (1,792 cores) 4432 images/sec. To the best of our knowledge, these are the best reported distributed memory training results for CPUs (in terms of efficiency). Previous work, which also uses GxM~\cite{sc18}, obtained on the same platform 1696 images/sec at the concurrency of 16 nodes, while our CNN kernels within GxM achieve 2239 images/sec, improving the end-to-end training performance by 1.32$\times$.  

\subsection{CNN kernels on integrated GPU and TVM}
\label{subsec:poc_cnn_results}
\begin{figure*}[t!]
\centering
\includegraphics[width=2.0\columnwidth]{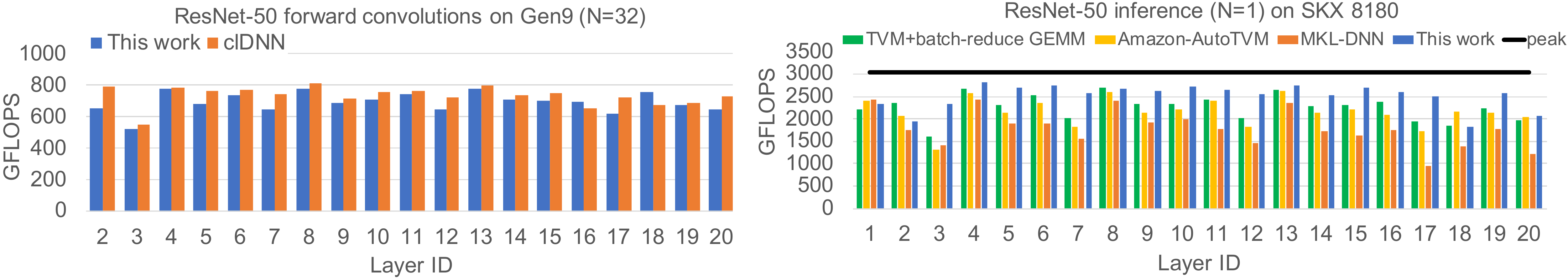}
\caption{CNN forward propagation: (Left) integrated GPU Gen9, (Right) implementation with batch-reduce GEMM in TVM}
\label{fig:poc_cnn}
\end{figure*}

In order to showcase the generalizability of our approach to diverse platforms, we developed the batch-reduce GEMM kernel in OpenCL. We implemented forward propagation CNN kernels (Algorithm~\ref{alg:cnn_br_gemm}) targeting Intel's integrated GPU Gen9 (Core i7 6770HQ) ~\cite{junkins2015compute} which has peak performance of 1152 GFLOPS. Figure~\ref{fig:poc_cnn}~(Left) illustrates the performance of our kernels (blue bars) and the vendor-optimized library Intel clDNN~\cite{cldnn} (orange bars). For the clDNN experiments, we tried all the available tensor layouts and picked the one that yields the highest performance. For this experiment the mini-batch size is $N$=32. We conclude that our kernels with batch-reduce GEMM achieve similar performance to clDNN. When considering the weighted efficiency, our kernels run at 728.3 Gflops and the clDNN kernels at 753.5 Gflops, thus our approach is within 3\% of the vendor-optimized, ad hoc implementation.

Once again, in this work, the specific algorithm/kernel development can be seen as loop tuning around batch-reduce GEMM, cf.\ section~\ref{sec:dl_algs}. Even though our DL kernels perform this tuning in a well-informed, manual fashion and are implemented in high-level C code, an alternative is to use a high-level tensor framework. Here we present results of such a proof-of-concept design, where we implement the forward convolutions within TVM~\cite{chen2018tvm}. More specifically, we provide to TVM the forward propagation loop recipes in high-level Python code, and at the innermost loop nest level we invoke our batch-reduce GEMM kernel.

To assess the efficiency of our approach, we consider the \emph{inference} use-case of CNNs (i.e.\ only the forward propagation pass). One idiosyncrasy of inference compared to training is the very small mini-batch $N$=1 that is necessitated to meet the latency requirements of the application. Figure~\ref{fig:poc_cnn}~(Right) shows the performance of our TVM implementation (green bars) on ResNet-50 forward kernels (mini-batch $N$=1) on the SKX 8180 platform. In this plot we also show the performance of the following implementations: i) CNN kernels within AutoTVM developed by Amazon~\cite{liu2018optimizing} (yellow bars), which are \emph{auto-tuned} for inference, ii) MKL-DNN (orange bars), and iii) the CNN kernel performance of our high-level C code kernels (blue bars). First, we observe that our DL kernels are the most efficient, achieving overall weighted efficiency of 2492 GFLOPS. Our Python implementation within TVM that exploits the batch-reduce GEMM kernel runs at 2361 GFLOPS, and consequently is within 5.3\% of our C implementation. Our TVM implementation is 2\% faster than the Amazon-AutoTVM \emph{auto-tuned code} and 1.24$\times$ faster than the vendor optimized MKL-DNN library. These results show that high-performance DL kernels within high-level tensor frameworks are feasible if the proper building block is used.

\section{Related Work}
The status quo in the development of high-performance DL workloads entails vendor-optimized DL primitives within some high-level deep learning framework (e.g. Tensorflow~\cite{tensorflow2015}, Pytorch~\cite{paszke2017pytorch}, Caffe~\cite{jia2014caffe}). MKL-DNN~\cite{mkldnn} is the Intel optimized DL library that provides specialized primitives (e.g.\ convolutions, RNN/LSTM cell, fully connected layers) for Intel CPUs. Each one of these primitives is individually optimized at a low-level on a per-platform basis in order to maximize performance, leading to numerous, highly specialized code-bases that do not generalize to different architectures. In an analogous way, cuDNN~\cite{chetlur2014cudnn} is the vendor-optimized DL library targeting Nvidia GPUs. clDNN~\cite{cldnn} is an open source performance library for DL applications intended for acceleration of DL Inference on Intel GPUs. All these library approaches suffer from the combinatorial explosion of the low-level optimizations that have to be done for each pair <architecture, DL primitive>. On the other hand, our proposed batch-reduce GEMM kernel covers all major DL primitives and it is the sole building-block that has to be optimized at a low-level.

An alternative methodology to implement DL primitives is to leverage vendor-optimized linear algebra library calls. For example, convolutions can be lowered to a matrix multiplication~\cite{chellapilla2006high}, but this lowering requires tensor transformations, and the obtained performance deteriorates~\cite{sc18,chetlur2014cudnn}. Aiming to accelerate the small matrix operations that pertain in DL, academia and industry have recently developed batched linear algebra routines ~\cite{dongarra2017design,stridedbatchgmemm,ng2017magmadnn}. The batched GEMM approaches in~\cite{stridedbatchgmemm,ng2017magmadnn} specifically target only Nvidia GPUs where the reduction across output subtensors is relatively cheap. Even though such approaches improve the performance of the DL primitives, they still perform worse than ad hoc implementations (cf.\ batched GEMM approach and mkl-dnn in Figure~\ref{fig:motivation}). Our work extends the batched GEMM routine, enables more optimizations (see Section~\ref{sec:br_gemm}), optimizes for locality and is therefore well suited for latency architectures such as CPUs. Also, the derived DL primitives match/exceed the performance of vendor-optimized ad hoc implementations as shown in Section~\ref{sec:results}.
 
Tensor compilers comprise a promising research area for end-to-end DL workload optimization and performance portability
(e.g.\ TVM~\cite{chen2018tvm}, GLOW~\cite{DBLP:journals/corr/abs-1805-00907}, PlaidML~\cite{plaidml}, MLIR~\cite{mlir}, Tensor Comprehensions~\cite{vasilache2018tensor}). Such frameworks treat tensors as first-class objects, and provide optimizations targeting tensor algebra programs (e.g.\ polyhedral optimizations for data movements). However, compilers struggle to optimize the GEMM-flavored loop nests for the nuances of the increasingly complex architectures. Our work can be seen as complementary to this effort, where the GEMM-flavored loops are abstracted into our batch-reduce GEMM call (which is independently optimized at a low-level). Then, the specific DL primitive optimization is reduced to mere loop tuning around a single kernel, and this task can be handed off to a tensor compiler. In Section~\ref{sec:results} we showcased a prototype for CNNs within TVM that uses our kernel.

\section{Conclusions}
In this work, we showed how the most popular DL algorithms (RNN/LSTM, CNN and MLP) can be formulated with batch-reduce GEMM as basic building block. We demonstrated that our methodology outperforms vendor-optimized, low-level DL primitives by factors up to 1.4$\times$. Moreover, we integrated our DL kernels into distributed frameworks, and optimized end-to-end workflows for GNMT and ResNet-50 training. In multi-node experiments we exceeded the performance of vendor-optimized implementations by up to 2.3$\times$. Additionally, we highlighted the architectural-agnostic aspect of our methodology by matching the CNN kernel performance of a vendor-provided library on integrated GPUs. Finally, we prototyped CNN kernels in a tensor compiler framework by harnessing our batch-reduce GEMM kernel, and matched the performance of \emph{auto-tuned} inference TVM primitives. As future work, we plan to extend our DL primitives for a wider set of architectures/workloads, and also we intend to experiment with Tensor compilers' automatic polyhedral optimization (e.g.~\cite{plaidml, poly}).

\FloatBarrier

\bibliographystyle{unsrt}
\bibliography{references}

\noindent Optimization Notice: Software and workloads used in
performance tests may have been optimized for performance only on
Intel microprocessors.  Performance tests, such as SYSmark and
MobileMark, are measured using specific computer systems,
components, software, operations and functions.  Any change to any
of those factors may cause the results to vary.  You should
consult other information and performance tests to assist you in
fully evaluating your contemplated purchases, including the
performance of that product when combined with other products.
For more information go to http://www.intel.com/performance.

\noindent Intel, Xeon, and Intel Xeon Phi are trademarks of Intel Corporation in the U.S. and/or other
countries.

\normalsize


\end{document}